\documentstyle[leqno,
	       epsf,
               lingmacros,
               fullname]{article}

\newdimen\lexentrydimen
\newdimen\lexcategorydimen
\newdimen\mothernodedimen
\def \lexentry #1#2#3{\setbox1\hbox{\it #1}\setbox2\hbox{#2}%
                      \ifdim \wd1 >3em \lexentrydimen=\wd1 %
                      \else \lexentrydimen=3em \fi%
                      \ifdim \wd2 >1em \lexcategorydimen=\wd2 %
                      \else \lexcategorydimen=1em \fi%
                      \noindent%
                      \makebox[\lexentrydimen][l]{\box1}\hspace*{1em}\nobreak%
                      \makebox[\lexcategorydimen][l]{\box2}\hspace*{1em}\nobreak%
                      \feqs{#3}}

\def \feqs #1{\begin{tabular}[t]{@{\strut}l@{\strut}}#1\end{tabular}}

\def \phraserule #1#2{\setbox1\hbox{#1}%
                      \ifdim \wd1 >2em \mothernodedimen=\wd1 %
                      \else \mothernodedimen=2em \fi%
   \leavevmode\hbox{\makebox[\mothernodedimen][l]{\box1}{$\longrightarrow$}\hspace{.6em}#2}}

\def \? {\leavevmode \llap {?}}
\def \up {\hbox {$\uparrow$\kern .2em}}
\def \down {\hbox {$\downarrow$\kern 0em}}

\def \obj {\hbox {\sc obj}}   
   
\def \obl {\hbox {\sc obl}}

\def \subj {\hbox {\sc subj}}

\def \pred {\hbox {\sc pred}}

\def \var {\hbox{\sc var}}

\def \restr {\hbox {\sc restr}}  
  
\def \spec {\hbox {\sc spec}}

\def \spec {\hbox {\sc spec}}

\makeatletter
\def\namecite{\leavevmode\def\citename##1{{##1}(}\@internalciteb}
\def\@citexb[#1]#2{\if@filesw\immediate\write\@auxout{\string\citation{#2}}\fi
  \def\@citea{}\@newcite{\@for\@citeb:=#2\do
    {\@citea\def\@citea{;\penalty\@m\ }\@ifundefined
       {b@\@citeb}{{\bf ?}\@warning
       {Citation `\@citeb' on page \thepage \space undefined}}%
\hbox{\csname b@\@citeb\endcsname}}}{#1}}
\def\@internalciteb{\@ifnextchar
[{\@tempswatrue\@citexb}{\@tempswafalse\@citexb[]}}
\def\@newcite#1#2{{#1\if@tempswa, #2\fi)}}

\def\downlett#1{$_{\hbox{\tiny\rm #1}}$}

\newcommand{\lam}[1]{\lambda #1. }
\newcommand{\linimp}{\;\mbox{$-\hspace*{-.4ex}\circ$}\;}
\newcommand{\means}{\makebox[1.2em]{$\leadsto$}}
\newcommand{\IT}[1]{\mbox{\it #1}\/}
\newcommand{\BF}[1]{\mbox{\bf #1}}
\newcommand{\intn}{\hat{\ }}
\newcommand{\extn}{\check{\ }}

\def\fd#1{\setlength{\baselineskip}{0pt}
  \small
     \hbox{#1}}
\def\fdx#1{\setlength{\baselineskip}{0pt}\vcenter{#1}}
\def\fdand#1{\(\left[\fdx{#1}\right]\)}

\def\feat#1#2{\vskip
.4ex\hbox{\hspace{.2em}#1\hspace{1em}#2\hspace{.2em}}\vskip .8ex}
\def\tightnode#1#2{\leavevmode\advance\nodemargin by -2pt
              \setbox1=\hbox{#2}\pscmd{/#1 \@int{\the\wd1} \space pt
                                           \@int{\the\ht1} \space pt
                                           \@int{\the\dp1} \space pt
                                       node}\box1\relax}
\let\@Hxfloat\@xfloat
\def\@xfloat#1[{\@ifnextchar{H}{\@HHfloat{#1}[}{\@Hxfloat{#1}[}}
\def\@HHfloat#1[H]{%
  \expandafter\let\csname end#1\endcsname\end@Hfloat
  \vskip\intextsep\vbox\bgroup\def\@captype{#1}\parindent\z@
  \ignorespaces}
\def\end@Hfloat{\egroup\vskip \intextsep}

\def\eqalign#1{\null\,\vcenter{\openup\jot\m@th
  \ialign{\strut\hfil$\displaystyle{##}$&$\displaystyle{{}##}$\hfil
      \crcr#1\crcr}}\,}
\def\fd#1{\setlength{\baselineskip}{0pt}
     \hbox{#1}}
\def\fdand#1{\(\left[\fdx{#1}\right]\)}
\def\fdx#1{\setlength{\baselineskip}{0pt}\vcenter{#1}}

\def\feat#1#2{\vskip
.4ex\hbox{\hspace{.2em}#1\hspace{1em}#2\hspace{.2em}}\vskip .8ex}

\def\line#1#2(#3){\@stepcounter{equation}
  \let\@currentlabel=\theequation \label{#3}
  \hbox to \hsize{\hbox to #1in{(\theequation) \hfil}
       \hskip 0pt minus 0.4in $#2$\hfil}}
\def\tightnode#1#2{\leavevmode\advance\nodemargin by -2pt
              \setbox1=\hbox{#2}\pscmd{/#1 \@int{\the\wd1} \space pt
                                           \@int{\the\ht1} \space pt
                                           \@int{\the\dp1} \space pt
                                       node}\box1\relax}

\makeatother

\newcommand{\pex}[1]{(\ref{#1})}

\newcommand{\All}[1]{\forall #1.\;}
\newcommand{\Al}{\IT{Al}\,}

\newcommand{\oneovermodb}[3]{\begin{array}[b]{@{}c@{}}
                                   #2 \\[-1.8ex]
                                   \hbox to #1{\hrulefill} \\[-.8ex]
                                   #3 \end{array}}

\newcommand{\oneover}[2]{\begin{array}{@{}c@{}} #1 \\
\hline #2  \end{array}}

\newcommand{\oneovermod}[3]{\begin{array}{@{}c@{}}
                                   #2 \\[-1.8ex]
                                   \hbox to #1{\hrulefill} \\[-.8ex]
                                   #3 \end{array}}

\title{Intensional Verbs Without Type-Raising or Lexical Ambiguity}
\author{
Mary Dalrymple\thanks{Xerox PARC, Palo Alto CA 94304;
{\tt \{dalrymple,lamping,saraswat\}@parc.xerox.com}}\\
John Lamping\footnotemark[1]\\
Fernando Pereira\thanks{AT\&T Bell Laboratories, Murray Hill NJ 07974;
{\tt pereira@research.att.com}}\\
Vijay Saraswat\footnotemark[1]}

\begin{document}
\maketitle

\section{Introduction}

We present an analysis of the semantic interpretation of intensional
verbs such as {\it seek}\/ that allows them to take direct objects of
either individual or quantifier type, producing both {\it de dicto}\/
and {\it de re}\/ readings in the quantifier case, all without needing
to stipulate type-raising or quantifying-in rules.  This simple
account follows directly from our use of logical deduction in
linear logic to express the relationship between syntactic structures
and meanings.  While our analysis resembles current categorial
approaches in important ways
(\cite{Moortgat:PhD,Moortgat:discontinuous,Morrill:type-logical,Carpenter:quant-scope}),
it differs from them in allowing the greater type flexibility of
categorial semantics (\cite{VanBenthem:LgInAction}) while maintaining
a precise connection to syntax. As a result, we are able to provide
derivations for certain readings of sentences with intensional verbs
and complex direct objects that are not derivable in current purely
categorial accounts of the syntax-semantics interface.  The analysis
forms a part of our ongoing work on semantic interpretation within the
framework of Lexical-Functional Grammar.

\section{Theoretical Background}

As a preliminary to presenting our analysis of intensional verbs,
we outline our approach to semantic interpretation in LFG.

\begin{figure}
\epsfxsize=4in
\centerline{\epsfbox{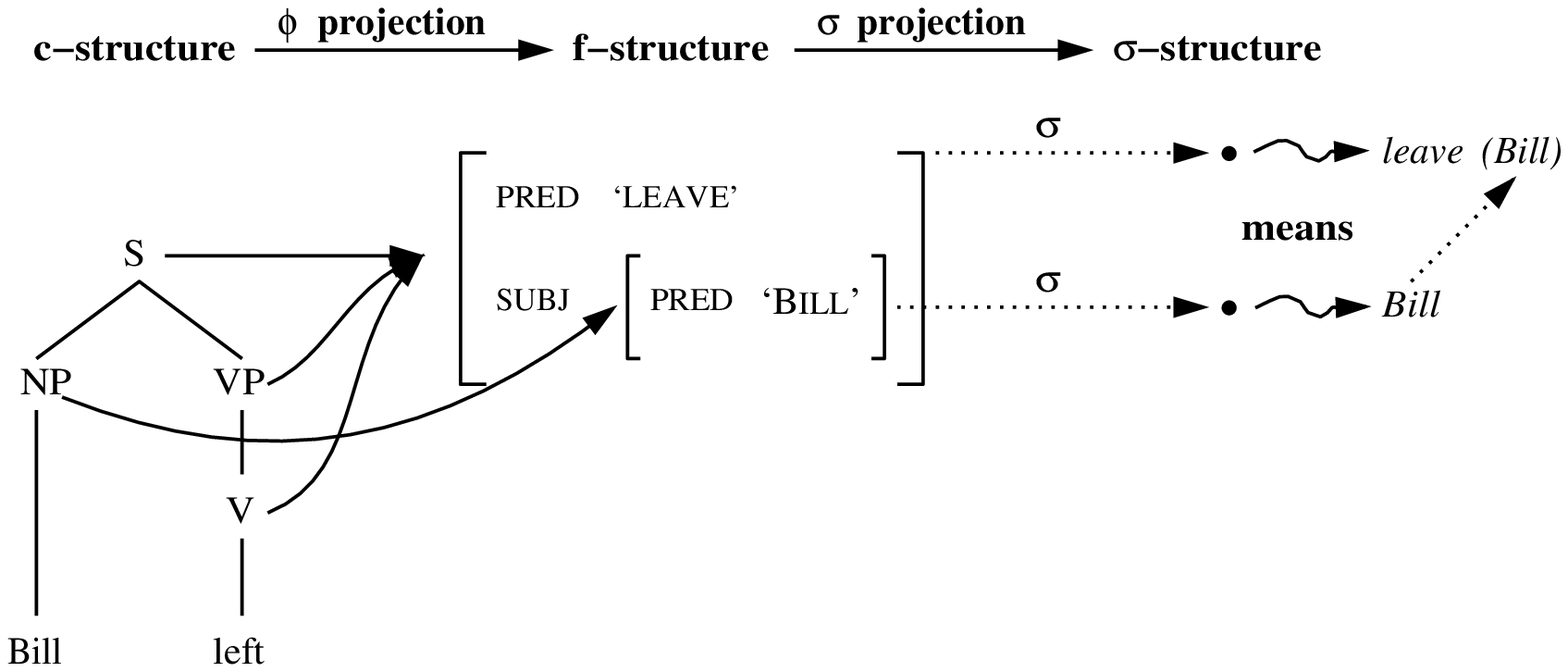}}
\caption{Semantic Interpretation Architecture}
\label{arch}
\end{figure}

It is well known that surface constituent structure does not always
provide the optimal set of constituents or hierarchical structure to
guide semantic interpretation.  This has led to efforts to develop
more abstract structures for the representation of relevant syntactic
information.  We follow Kaplan and Bresnan (1982)
\nocite{KaplanBresnan:LFG} and \namecite{Halvorsen:LI} in taking the
functional structure or {\it f-structure}\/ of LFG as the primary
input to semantic interpretation. The syntactic structures of LFG, the
constituent structure or {\it c-structure}\/ and the f-structure, are
related by means of a functional correspondence, represented in Figure
\ref{arch} by solid lines leading from nodes of the c-structure tree
to f-structures (\cite{KaplanBresnan:LFG}).\footnote{The c-structure
and f-structure presented here have been simplified to show only the
detail necessary for the semantic issues addressed here.  We also do
not address a number of orthogonal semantic issues (tense and aspect,
for example), providing only enough details of the representation of
the meaning of a sentence to illustrate the points relevant to the
discussion at hand.}
In more recent work, \namecite{Kaplan:3Sed} and
\namecite{HalvorsenKaplan:Projections} have proposed to extend
the theory of correspondences to other structures, called {\it
projections}\/.  Here, we will appeal to a semantic projection
$\sigma$, relating f-structures and their meanings. Notationally, a
subscript $\sigma$ will indicate the semantic or $\sigma$ projection
of an f-structure $f$, so that the semantic projection of $f$ will be
written $f_\sigma$.  In Figure \ref{arch}, dotted lines represent the
relation between f-structures and their semantic projections.
Finally, as shown in the figure,
the semantic projection $f_\sigma$ of an f-structure $f$ can be put in
correspondence with a meaning $\phi$:
\enumsentence{$f_\sigma \means \phi$}
Informally, we read this expression as ``the meaning of $f$ is
$\phi$''.  We use expressions of this sort to lexically associate
meanings with f-structures, as in the following lexical entry
for the word {\it Bill}\/:
\enumsentence{\label{bill-entry}\begin{tabular}[t]{ll}
Bill & ($\uparrow$ {\sc pred}) = {\sc `Bill'}\\
     & $\uparrow_\sigma \means \IT{Bill}$
\end{tabular}}
The first line of this lexical entry:
\begin{center}
($\uparrow$ {\sc pred}) = {\sc `Bill'}
\end{center}
says (roughly) that the word {\it Bill}\/ introduces an f-structure $\uparrow$
whose \pred\ is `Bill'.  The second line:
\begin{center}
$\uparrow_\sigma \means \IT{Bill}$
\end{center}
says that the meaning of that f-structure is $\IT{Bill}$.  When a
lexical entry is used, the metavariable `$\uparrow$' is instantiated
with some constant $f$ denoting an f-structure
(\cite[page~183]{KaplanBresnan:LFG}).  In particular, the term
$\uparrow_\sigma$ is instantiated to the semantic projection
$f_{\sigma}$ of $f$ and the formula \hbox{$\uparrow_\sigma \means
\IT{Bill}$} is instantiated to $f_{\sigma}
\means \IT{Bill}$, asserting that the meaning of $f$ is $\IT{Bill}$.

More complicated lexical entries give not only meanings for
f-structures, but also instructions for assembling f-structure
meanings from the meanings of other f-structures.  We distinguish a
{\em meaning language}, in which we represent the meanings of
f-structures, and a {\em composition language} or {\it glue
language}\/, in which we describe how to assemble the meanings of
f-structures from the meanings of their substructures. Each lexical
entry will contain a composition language formula, its {\em meaning
constructor}, specifying how a lexical entry contributes to the
meaning of any structure in which it participates.

In principle, the meaning language can be any suitable logic. Here,
since we are concerned with the semantics of intensional verbs, we
will use Montague's higher-order intensional logic IL.  The
expressions that appear on the right side of the $\means$ connective
in lexical entries like \pex{bill-entry} above are (usually open)
terms in the meaning language.

Our composition language is a fragment of linear logic with
higher-order quantification. While the resource sensitivity of linear
logic is crucial to our overall interpretation framework, it does not
play a central role in the analysis discussed here, so the linear
connectives can be informally read as their classical
counterparts.\footnote{We make use of {\em linear logic}
(\cite{Girard:Linear}) because its resource sensitivity turns out to
be a good match to natural language.  This property of linear logic
nicely captures, among other things, the LFG requirements of {\it
coherence}\/ and {\it consistency}\/, and enables a nice treatment of
modification (\cite{DLS:EACL}), quantification
(\cite{DLPS:quant-lfg}), and complex predicates
(\cite{DHLS:ROCLING}). We make use only of the {\it tensor fragment}\/
of linear logic. The fragment is closed under conjunction, universal
quantification and implication. It arises from transferring to linear
logic the ideas underlying the concurrent constraint programming
scheme of Saraswat (\shortcite{Saraswat:PhD}). A nice tutorial
introduction to linear logic itself may be found in Scedrov
(\shortcite{Scedrov:Linear}).} In contrast to standard approaches, which
use the $\lambda$-calculus to combine fragments of meaning via ordered
applications, we combine fragments of meaning through unordered
conjunction and implication.  Rather than using $\lambda$-reduction to
simplify meanings, we rely on deduction, as advocated by Pereira
(\shortcite{Pereira:SemComp,Pereira:HOD}).

\section{A Simple Example}

We now turn to a simple example to illustrate the framework.  The
lexical entries necessary for the example in Figure \ref{arch}
are:\footnote{ In the composition language, we use upper-case letters
for {\em essentially existential} variables, that is, Prolog-like
variables that become instantiated to particular terms during a
derivation, and lower-case letters for {\em essentially universal
variables} that stand for new local constants (also called
eigenvariables) during a derivation.}

\enumsentence{
\begin{tabular}[t]{ll}
Bill & ($\uparrow$ {\sc pred}) = {\sc `Bill'}\\
     & $\uparrow_\sigma$ = $\IT{Bill}$\\[1ex]
left & ($\uparrow$ {\sc pred})= `{\sc leave}'\\
     & $\forall X.\ (\uparrow \mbox{\sc subj})_\sigma \means X \linimp
    \uparrow_\sigma \means \IT{leave}(X)$
\end{tabular}}
The symbol `$\linimp$' is the {\it linear implication}\/ operator of linear
logic; for this paper, `$\linimp$' can be thought of as analogous to
classical implication `$\rightarrow$'.  The formula $$\forall X.\
(\uparrow \mbox{\sc subj})_\sigma
\means X \linimp \uparrow_\sigma \means \IT{leave}(X)$$ states that
the verb {\it left}\/ requires a meaning $X$ for its subject,
$(\uparrow \mbox{\sc subj})$; when that meaning is provided, the
meaning for the sentence will be $\IT{leave}(X)$.  When the words {\it
Bill}\/ and {\it left}\/ are used in a sentence, the metavariable
$\uparrow$ will be instantiated to a particular f-structure, and the
meaning given in the lexical entry will be used as the meaning of that
f-structure.

Here we repeat the f-structure in Figure \ref{arch}, including labels
for ease of reference:
\enumsentence{\evnup{
\fd{$f$:\fdand{\feat{\pred}{\sc `leave'}
           \feat{\subj}{$g$:\fdand{\feat{\pred}{\sc `Bill'}}}}}}}
Annotated phrase structure rules provide instructions for assembling
this f-structure by instantiating the metavariables `$\uparrow$' in
the lexical entries above.  For instance, the metavariable
`$\uparrow$' in the lexical entry for {\it Bill}\/ is instantiated to
the f-structure labeled $g$.

{}From the instantiated lexical entries of {\it Bill}\/ and {\it left}\/, we
have the following semantic information:
\enumsentence{$\begin{array}[t]{ll}
\BF{leave}\colon & \forall X.\ g_{\sigma} \means X \linimp f_{\sigma} \means
\IT{leave}(X)\\
\BF{Bill}\colon & g_{\sigma} \means \IT{Bill}
\end{array}$}
where {\bf leave} and {\bf Bill} are names for their respective
formulas.  By modus ponens, we deduce:
$$ \BF{Bill}, \BF{leave} \vdash f_\sigma\means \IT{leave}(\IT{Bill})$$

The elements of the f-structure provide a set of formulas in the
composition logic that introduce semantic elements and describe how
they can be combined.  For example, lexical items for words that
expect arguments, like verbs, typically contribute a formula for
combining the meanings of their arguments into a result.  Once all the
formulas are assembled, deduction in the logic is used to infer the
meaning of the entire structure.  Throughout this process we maintain
a clear distinction between meanings proper and assertions about
meaning combinations.

\section{Quantification}

We now turn to an overview of our analysis of quantification
(Dalrymple, Lamping, Pereira, and Saraswat 1993).  As a simple
example, consider the sentence
\enumsentence{
Every man left.}
For conciseness, we will not illustrate the combination of the meaning
constructors for {\it every}\/ and {\it man}; instead, we will work
with the derived meaning constructor for the subject {\it every
man}\/, showing how it combines with the meaning constructor for {\it
left}\/ to produce a meaning constructor giving the meaning of the
whole sentence.

The basic idea of our analysis of quantified NPs can be seen as a
logical counterpart at the semantic composition level of the
generalized-quantifier type assignment for (quantified) NPs
(\cite{Barwise:Generalized}). Under that assignment, a NP meaning $Q$
has type $$(e\rightarrow t)\rightarrow t$$---that is, a function from
a property, the scope of quantification, to a proposition.  At the
semantic composition level, we can understand that type as follows. If
by assuming that $x$ is the entity referred to by the NP we can derive
$S x$ as the meaning of the scope of quantification, where $S$ is a
property (a function from entities to propositions), then we can
derive $Q S$ as the meaning of the whole clause containing the NP.

The f-structure for the sentence {\it Every man left} is:
\enumsentence{\evnup{
\fd{$f$:\fdand{\feat{\pred}{\sc `leave'}
           \feat{\subj}{$g$:\fdand{\feat{\spec}{\sc `every'}
                                     \feat{\pred}{\sc `man'}}}}}}}
The meaning constructors for {\it every man}\/ and {\it left}\/
are:\footnote{We use throughout the convenient abbreviation
$Q(x,Rx,Sx)$ for the application of the generalized quantifier $Q$ to
restriction $R$ and scope $S$.}

\enumsentence{\label{ex:every-man-premises}
$\begin{array}[t]{lr@{\ }l}
\BF{leave}\colon & \forall X. & \/g_{\sigma} \means X
\linimp f_{\sigma} \means \IT{leave}(X)\\[2ex]
\BF{every-man}\colon\ & \forall H, S. &
(\forall x.  g_{\sigma} \means x \linimp H \means Sx)\\
&& \linimp H \means \IT{every}(z, \IT{man}(z), Sz)
\end{array}$}
The meaning constructor for {\it left\/} is as before.  The meaning
constructor for {\it every
man\/} quantifies over semantic projections $H$ which constitute possible
quantification
scopes; its propositional structure corresponds to the
standard type assignment, $(e\rightarrow t)\rightarrow t $.  It can be
paraphrased as:
\[
\begin{array}{ll}
\All{H,S} \\
\quad (\All{x}g_{\sigma} \means x & \left\{ \parbox{30ex}{if, by
assuming an arbitrary meaning $x$ for $g$,} \right. \\[2.5ex]
\qquad \linimp H \means Sx) & \left\{ \parbox{30ex}{a meaning $Sx$ for some
scope $H$ can
be derived,} \right. \\[2.5ex]
\quad\ \linimp {H} \means \IT{every}(z, \IT{man}(z), Sz) & \left\{
\parbox{30ex}{then a possible complete meaning for $H$ can be derived.}\right.
\end{array}
\]
\noindent In the case at hand, the semantic projection $f_\sigma$ will be
chosen to provide the scope of quantification.\footnote{From what has
been said so far, $g_\sigma$ could also be chosen to
provide the scope, leading to a nonsensical result.  As explained in
our full analysis of quantifiers (Dalrymple, Lamping, Pereira, and
Saraswat 1993), that problem is avoided by using a family of $\means$
relations indexed by the type of their second argument.  The relation
for the meaning of the scope of quantification is the one that expects
a proposition meaning, so $g_\sigma$ can not provide
a scope.}  It has exactly the form that the antecedent of {\bf
every-man} expects.  The meaning $S$ will be instantiated to $\lambda
x. \IT{leave}(x)$.  From the premises in
(\ref{ex:every-man-premises}), we can deduce the meaning of the scope
f-structure $f$:
$$\BF{every-man}, \BF{leave} \vdash f_\sigma\means every\/(z,
man\/(z), \IT{leave}\/(z))$$
The resource sensitivity of linear logic ensures that the scope of
quantification is constructed and used exactly once.

\section{Intensional Verbs}

We follow Montague (\shortcite{Montague:PTQ}) in requiring intensional
verbs like {\it seek\/} to take an object of NP type.  What is
interesting is that this is the only step required in our setting to
obtain the appropriate ambiguity predictions for intensional verbs.
The {\it de re\/}/{\it de dicto\/} ambiguity of a sentence like {\it
Bill seeks a unicorn}\/:
\[
\begin{array}{ll}
\mbox{{\em de dicto}\/ reading:} & \IT{seek}\/(\IT{Bill}, {\intn}\lambda Q.
a(x, \IT{unicorn}(x), [{\extn}Q](x)))\\
\mbox{{\em de re}\/ reading:} & \IT{a}(x, \IT{unicorn}(x),
\IT{seek}\/(\IT{Bill}, {\intn}\lambda Q. [{\extn}Q](x)))
\end{array}
\]
\noindent is a natural consequence, in our setting, of {\it
seek} taking an NP-type argument.

We assign the following lexical entry to the verb {\it seek}\/:

\enumsentence{
\begin{tabular}[t]{l@{\hspace{2pt}}l@{}r@{}l}
seek & \multicolumn{3}{l}
       {($\uparrow$ {\sc pred}) = {\sc `seek'}}\\
     & $\forall Z, Y.$&&$(\uparrow \subj)_\sigma \means Z$\\
     &                &$\otimes$&$(\forall s, p. (\forall X.
   (\uparrow \obj)_\sigma \means X \linimp s \means p(X))
   \linimp s \means Y({\intn}p))$\\
   &\multicolumn{3}{l}{\quad$\linimp\uparrow_\sigma \means \IT{seek}(Z,
{\intn}Y)$}
\end{tabular}}

\noindent The significant fact here is that {\it seek}\/ differs from
an extensional verb such as {\it find}\/ below (corresponding to the type
$e\rightarrow e\rightarrow t$) in its specification of requirements on
its object:
\enumsentence{\begin{tabular}[t]{l@{\hspace{2pt}}l@{}r@{}l}
find & \multicolumn{3}{l}
       {($\uparrow$ {\sc pred}) = {\sc `find'}}\\
     & $\forall Z, Y.(\uparrow \subj)_\sigma \means Z\otimes
(\uparrow \obj)_\sigma\means Y \linimp\uparrow_\sigma \means \IT{find}(Z,Y)$
\end{tabular}}
Note also the use of the operators ``$\intn$'' and ``$\extn$'' of
IL.  Computationally, this implies that our proofs have to be carried
out in a logic whose terms are (typed) lambda-expressions that satisfy
$\alpha-$, $\beta-$ and $\eta$-equality and also the law
${\extn}({\intn}P)=P$, for all $P$.

The lexical entry for {\it seek}\/ can be paraphrased as follows:
\[
\begin{array}[t]{ll}
\All{Z, Y}(\uparrow \subj)_\sigma \means Z \otimes &
\left\{\parbox{30ex}{The verb {\it seek}\/ requires a meaning $Z$ for its
subject and}\right. \\[2.5ex]
\begin{array}{ll@{\mbox{}}l}\lefteqn{(\forall s, p.} \\
&  \quad (\All{X} & (\uparrow \obj)_\sigma \means X \\
& & \linimp s \means p(X)) \\
&  \multicolumn{2}{l}{\linimp s \means Y({\intn}p))}
\end{array}
&\left\{\parbox{30ex}{a meaning ${\intn}Y$ for its object,
where $Y$ is an NP meaning
applied to the meaning $p$ of an
arbitrarily-chosen `scope' $s$,}\right.\quad(*) \\[8ex]
\quad\linimp\uparrow_\sigma \means \IT{seek}(Z, {\intn}Y) &
\left\{\parbox{30ex}{to produce the clause meaning $\IT{seek}(Z,
{\intn}Y)$.}
\right.
\end{array}
\]

\noindent Rather than looking for an entity type meaning for its
object, the requirement expressed by the subformula labeled $(*)$
exactly describes the form of quantified NP meanings discussed in the
previous section.  In this case, a quantified NP in the object
position is one that can accept anything that takes a meaning for
$(\uparrow \obj)_\sigma$ to a meaning for any $s$, and convert that
into a meaning for the $s$.  In particular, the quantified NP {\it a
unicorn}\/ will fill the requirement, as we now demonstrate.

The f-structure for {\it Bill seeks a unicorn}\/ is:
\enumsentence{\evnup{
\fd{$f$:\fdand{\feat{\pred}{\sc `seek'}
           \feat{\subj}{$g$:\fdand{\feat{\pred}{\sc `Bill'}}}
           \feat{\obj}{$h$:\fdand{\feat{\spec}{\sc `a'}
                                    \feat{\pred}{\sc `unicorn'}}}}}}}
The semantic information associated with this f-structure is:
$$\begin{array}[t]{l@{\hspace{2pt}}l}
\BF{seeks}\colon & \forall Z, Y.
   \begin{array}[t]{r@{}l}&g_\sigma \means Z \\
                   \otimes&(\forall s, p. (\forall X.
      h_\sigma \means X \linimp s \means p(X))
      \linimp s \means Y({\intn}p))
   \end{array} \\
   &\quad \linimp f_\sigma \means \IT{seek}(z, {\intn}Y) \\
\BF{Bill}\colon & g_{\sigma} \means \IT{Bill}\\
\BF{a-unicorn}\colon\ & \forall {H}, S.
\/(\forall x.  h_{\sigma} \means x \linimp {H} \means Sx)
\linimp {H} \means a\/(z, \IT{unicorn}\/(z), Sz)
\end{array}$$
These are the premises for the deduction of the meaning of the
sentence {\it Bill seeks a unicorn\/}.  From the premises {\bf Bill}
and {\bf seeks}, we can conclude by modus ponens:
$$\begin{array}[t]{ll}
\BF{Bill-seeks}\colon & \forall Y. (\forall s, p. (\forall X.
   h_\sigma \means X \linimp s \means p(X))
   \linimp s \means Y({\intn}p))\\
   & \qquad \linimp {f_\sigma} \means \IT{seek}(\IT{Bill}, {\intn}Y)
\end{array}$$
Different derivations starting from the premises
\BF{Bill-seeks} and \BF{a-unicorn} will yield the different readings
of {\em Bill seeks a unicorn\/} that we seek.
\subsection{De Dicto Reading}
The formula {\bf a-unicorn}
is exactly what is required by the antecedent
of {\bf Bill-seeks} provided that the following substitutions are
performed:
\[\eqalign{H &\mapsto s \cr
S & \mapsto p \cr
X & \mapsto x \cr
Y & \mapsto \lambda P. \IT{a}(z,
\IT{unicorn}(z), [{\extn}P](z))\cr
}\]
We can thus conclude the desired {\it de dicto}\/ reading:
$$f_\sigma \means \IT{seek}(\IT{Bill}, {\intn}\lambda P. \IT{a}(z,
\IT{unicorn}(z), [{\extn}P](z)))$$

To show how the premises also support a {\it de re} reading, we take
first a short detour through a simpler example.

\subsection{Nonquantified Objects}

The meaning constructor for {\it seek} also allows for
nonquantified objects as arguments, without needing a special
type-raising rule.  Consider the f-structure for the sentence {\it
Bill seeks Al}:

\enumsentence{\evnup{
\fd{$f$:\fdand{\feat{\pred}{\sc `seek'}
           \feat{\subj}{$g$:\fdand{\feat{\pred}{\sc `Bill'}}}
           \feat{\obj}{$h$:\fdand{\feat{\pred}{\sc `Al'}}}}}}}

\noindent The lexical entry for {\it Al}\/ is analogous to the one for
{\it Bill}\/.  We begin with the premises {\bf Bill-seeks} and {\bf
Al}:
$$\begin{array}[t]{ll}
\BF{Bill-seeks}\colon & \forall Y. (\forall s, p. (\forall X.
   h_\sigma \means X \linimp s \means p(X))
   \linimp s \means Y({\intn}p))\\
   & \qquad \linimp {f_\sigma} \means \IT{seek}(\IT{Bill}, {\intn}Y) \\
\BF{Al}\colon\ & h_{\sigma} \means \Al
\end{array}$$
For the derivation to proceed, {\bf Al} must
supply the NP meaning constructor that {\bf
Bill-seeks} requires.  This is possible because {\bf Al} can map a proof $\Pi$
of the meaning for $s$ from the meaning for $h$ into a meaning
for $s$, simply by supplying $h_{\sigma} \means\Al$ to $\Pi$.
\begin{figure}
$$
\oneover{h_\sigma \means \Al \vdash h_\sigma \means \Al \qquad
s \means P(\Al) \vdash  s \means P(\Al)}
{
\oneover{h_\sigma \means \Al, h_\sigma \means \Al
\linimp s \means P(\Al) \vdash  s \means P(\Al)}
{
\oneover{h_\sigma \means \Al, (\forall x. h_\sigma \means x
\linimp s \means P(x)) \vdash  s \means P(\Al)}
{
\oneover{h_\sigma \means \Al \vdash (\forall x. h_\sigma
\means x \linimp s \means P(x)) \linimp  s \means
P(\Al)}
{h_\sigma \means \Al
\vdash
\forall P. (\forall x. h_\sigma
\means x \linimp s \means P(x)) \linimp  s \means
P(\Al)}}}}\\
$$
\caption{Proof that {\bf Al} can function as a quantifier}
\label{Al-proof}
\end{figure}
Formally, from {\bf Al} we can prove (Figure~\ref{Al-proof}):
\enumsentence{\label{type-raised-al}$\forall P. (\forall x.  h_\sigma \means x
\linimp s \means P(x)) \linimp s \means P(\Al)$}
This corresponds to the
Montagovian type-raising of a proper name meaning to an NP meaning,
and also to the undirected Lambek calculus derivation of
the sequent $e\Rightarrow(e\rightarrow t)\rightarrow t$.

Formula \pex{type-raised-al} with the substitutions
\[P \mapsto p, Y \mapsto \lambda P. [{\extn}P](\Al)
\]
can then be used to satisfy the
antecedent of {\bf
Bill-seeks} to yield the desired result:
$$f_\sigma \means \IT{seek}(\IT{Bill}, {\intn}\lambda P.  [{\extn}P](\Al))$$

It is worth contrasting the foregoing derivation with treatments of
the same issue in the lambda calculus.  The function $\lambda
x. \lambda P. Px$ raises a term like $\Al$ to the quantified NP
form $\lambda P. P(\Al)$, so it is easy to modify $\Al$ to
make it suitable for {\bf seek}.  But the conversion must be
explicitly applied somewhere, either in a meaning postulate or in an
alternate definition for {\em seek}, in order to carry out the
derivation.  This is because a lambda calculus function must specify
exactly what is to be done with its arguments and how they will
interact.  It must presume some functional form of its arguments in
order to achieve its own function.  Thus, it is impossible to write a
function that is indifferent with respect to whether its argument is
{\em Al} or $\lambda P. P(\Al)$.

In the deductive framework, on the other hand, the exact way in which
different propositions can interact is not fixed, although it is
constrained by their (logical and quantificational) propositional
structure.  Thus \mbox{$h_{\sigma}
\means \Al$} can function as any logical consequence of itself,
in particular as: $$\All{S,P}(\All{x}h_\sigma \means
x \linimp S \means P(x)) \linimp S \means P(\Al)$$
This flexibility, which is also found in syntactic-semantic analyses
based on the Lambek calculus and its variants
(\cite{Moortgat:PhD,Moortgat:Labelled,VanBenthem:LgInAction}), seems
to align well with some of the type flexibility in natural language.

\begin{figure}
$$
\oneover{
\oneover{
\oneover{
\oneover{
\oneover{
\oneover{I \means Z\vdash I \means Z \qquad
S \means P(Z) \vdash  S \means P(Z)}
{I \means Z, I \means Z
\linimp S \means P(Z) \vdash  S \means P(Z)}}
{I \means Z, (\All{x} I \means x
\linimp S \means P(x)) \vdash  S \means P(Z)}}
{I \means Z \vdash (\All{x} I
\means x \linimp S \means P(x)) \linimp  S \means
P(Z)}}
{I \means Z
\vdash
\All{S,P} (\All{x}I
\means x \linimp S \means P(x)) \linimp  S \means
P(Z)}}
{\vdash I \means Z \linimp
\All{S,P} (\All{x}I
\means x \linimp S \means P(x)) \linimp  S \means
P(Z)}}
{\vdash \All{I,Z} I \means Z \linimp
\All{S,P} (\All{x}I
\means x \linimp S \means P(x)) \linimp  S \means
P(Z)}
$$
\caption{General Type-Raising Theorem}
\label{type-raising-proof}
\end{figure}

\subsection{Type Raising and Quantifying In}
The derivation in Figure \ref{Al-proof} can be generalized as shown in
Figure \ref{type-raising-proof} to produce the general type-raising
theorem:
\enumsentence{\label{type-raising-theorem}
$\All{I,Z} I \means Z \linimp (\All{S,P} (\forall x. I
\means x \linimp S \means P(x)) \linimp S\means P(Z))$
}
This theorem can be used to raise meanings of $e$ type to
$(e\rightarrow t)\rightarrow t$ type, or, dually, to quantify into
verb argument positions. For example, with the variable instantiations
\[\eqalign{I &\mapsto h_\sigma\cr
X &\mapsto  x\cr
P &\mapsto p\cr
S &\mapsto s\cr
Y &\mapsto \lambda R. [{\extn}R](Z)\cr}
\]
we can use transitivity of implication to combine
\pex{type-raising-theorem} with {\bf Bill-seeks} to derive:
\[
\BF{Bill-seeks}'\colon \All{Z}h_\sigma \means Z \linimp f_\sigma
\means \IT{seek}(\IT{Bill}, {\intn}\lambda R.  [{\extn}R](Z))
\]
This formula can then be combined with arguments of type $e$ to
produce a meaning for $f_\sigma$. For instance, it will take the
non-type-raised $h_\sigma \means \Al$ to yield the same result
\[f_\sigma \means\IT{seek}(\IT{Bill}, {\intn}\lambda R.  [{\extn}R](\Al))\]
as the combination of {\bf Bill-seeks} with the type-raised version of \BF{Al}.
In fact, $\BF{Bill-seeks}'$ corresponds to type $e\rightarrow t$, and can
thus be used as the scope of a quantifier, which would then quantify
into the intensional direct object argument of {\em seek}. As we will
presently see, that is exactly what is needed to derive
{\em
de re} readings.
\subsection{De Re Reading}

We have just seen how theorem \pex{type-raising-theorem} provides a
general mechanism for quantifying into intensional argument
positions. In particular, it allowed the derivation of $\BF{Bill-seeks}'$
from \BF{Bill-seeks}. Now, given the premises
$$\begin{array}[t]{@{}ll@{}}
\BF{Bill-seeks}'\colon & \All{Z}h_\sigma \means Z \linimp f_\sigma
\means \IT{seek}(\IT{Bill}, {\intn}\lambda R.  [{\extn}R](Z)) \\
\BF{a-unicorn}\colon\ & \All{H, S}(\All{x}
h_{\sigma} \means x \linimp H \means Sx)
\linimp H \means \IT{a}(z, \IT{unicorn}(z), Sz)
\end{array}$$
and the variable substitutions
\[\eqalign{Z & \mapsto x \cr
H & \mapsto f_\sigma \cr
S & \mapsto \lambda z. \IT{seek}(\IT{Bill}, {\intn}\lambda R.
[{\extn}R](z))\cr}
\]
we can apply modus ponens to derive the {\em de re} reading of {\em Bill seeks
a unicorn}\/:
\[
f_\sigma \means \IT{a}(z, \IT{unicorn}(z),
\IT{seek}(\IT{Bill}, {\intn}\lambda R.  [{\extn}R](z)))
\]

\subsection{A Comparison with Categorial Approaches}
The analysis presented here has strong similarities to analyses of the
same phenomena discussed by \namecite{Morrill:type-logical} and
\namecite{Carpenter:quant-scope}. Following
\namecite{Moortgat:discontinuous}, they add to an appropriate version of
the Lambek calculus (\cite{Lambek:sentstruct}) the {\em scope}
connective $\Uparrow$, subject to the following proof rules:
\[
\begin{array}{c}
\oneover{\Gamma, v:A, \Gamma' \Rightarrow u:B\qquad \Delta, t(\lam{v}u):B,
\Delta' \Rightarrow C}
{\Delta, \Gamma, t:A\Uparrow B, \Gamma', \Delta' \Rightarrow
C}\quad\mbox{[QL]} \\[1em]
\oneover{\Gamma \Rightarrow u:A}
{\Gamma \Rightarrow \lam{v}v(u):A\Uparrow B}\quad\mbox{[QR]}
\end{array}
\]
\noindent In terms of the scope connective, a quantified noun phrase is given
the category $\mbox{N}\Uparrow\mbox{S}$, which semantically
corresponds to the type \mbox{$(e\rightarrow t)\rightarrow t$} and
agrees with the propositional structure of our linear formulas for
quantified noun phrases, for instance (\ref{ex:every-man-premises}). A
phrase of category $N\Uparrow S$ is an infix functor that binds a
variable of type $e$, the type of individual noun phrases N, within a
scope of type $t$, the type of sentences S.  An intensional verb like
`seek' has, then, category
$(\mbox{N}\setminus(\mbox{S})/(\mbox{N}\Uparrow\mbox{S})$, with
corresponding type $(((e\rightarrow t)\rightarrow t)\rightarrow
e\rightarrow t)$. \footnote{These category and type assignments are an
oversimplification since intensional verbs like `seek' require a
direct object of type $s\rightarrow ((e\rightarrow t)\rightarrow t)$,
but for the present discussion the simpler category and type are
sufficient.
\namecite{Morrill:type-logical} provides a full treatment.} Thus the
intensional
verb will take as direct object a quantified noun phrase, as required.

A problem arises, however, with sentences such as
\enumsentence{Bill seeks a conversation with every unicorn.\label{conv}}
This sentence has five possible interpretations:
\eenumsentence{
\item $\IT{seek}(\IT{Bill}, {\intn}\lambda P.
\IT{every}(u,\IT{unicorn}\/(u),a\/(z,\IT{conv-with}(z,u),[{\extn}P](z))))$\label{conv-2}
\item $\IT{seek}(\IT{Bill}, {\intn}\lambda P. a\/(z,
\IT{every}(u,\IT{unicorn}\/(u),\IT{conv-with}(z,u)),[{\extn}P](z)))$\label{conv-1}
\item $\IT{every}(u,\IT{unicorn}\/(u), \IT{seek}(\IT{Bill}, {\intn}\lambda P.
a\/(z,\IT{conv-with}(z,u),[{\extn}P](z))))$\label{conv-3}
\item $\IT{every}(u,\IT{unicorn}\/(u),
a\/(z,\IT{conv-with}(z,u),
\IT{seek}(\IT{Bill}, {\intn}\lambda P. [{\extn}P](z))))\label{conv-4}$
\item $\IT{a}\/(z,
\IT{every}(u,\IT{unicorn}\/(u),\IT{conv-with}(z,u)),
\IT{seek}(\IT{Bill}, {\intn}\lambda P. [{\extn}P](z)))\label{conv-5}$
}

\noindent Both our approach and the categorial analysis using the scope
connective have no problem in deriving interpretations (\ref{conv-1}),
(\ref{conv-3}), (\ref{conv-4}) and (\ref{conv-5}). In those cases, the
scope of `every unicorn' is interpreted as an appropriate term of type
$e\rightarrow t$. However, the situation is different for
interpretation (\ref{conv-2}), in which both the conversations and the
unicorn are {\em de dicto}, but the conversations sought may be
different for different unicorns sought. As we will show below, this
interpretation can be easily derived within our framework. However, a
similar derivation does not appear possible in terms of the categorial
scoping connective.

The difficulty for the categorial account is that the category
$\mbox{N}\Uparrow\mbox{S}$ represents a phrase that plays the role of
a category N phrase where it appears, but takes an S (dependent on the
N) as its scope. In the derivation of (\ref{conv-2}), however, the
scope of `every unicorn' is `a conversation with', which is not of
category S.  Semantically, `a conversation with' is represented
by:
\enumsentence{$
\lambda P. \lambda
u. a\/(z,\IT{conv-with}(z,u),[{\extn}P](z)):(e\rightarrow
t)\rightarrow (e \rightarrow t)$\label{a-conv-type}}
The {\em undirected} Lambek calculus (\cite{VanBenthem:LgInAction})
allows us to compose (\ref{a-conv-type}) with the interpretation of
`every unicorn':
\enumsentence{$\lambda Q.
\IT{every}(u,\IT{unicorn}\/(u), Q(u)):(e\rightarrow t)\rightarrow
t$\label{e-u-typ}}
to yield:
\enumsentence{$\lambda
P.
\IT{every}(u,\IT{unicorn}\/(u),a\/(z,\IT{conv-with}(z,u),[{\extn}P](z))):(e\rightarrow
t)\rightarrow t$\label{e-u-a-c-typ}}
As we will see below, our linear logic formulation also allows that
derivation step.

In contrast, as \namecite{Moortgat:discontinuous} points out, the
categorial rule [QR] is not powerful enough to raise $N\Uparrow S$ to
take as scope any functor whose result is a S. In particular, the
sequent
\enumsentence{$\mbox{N}\Uparrow \mbox{S}\Rightarrow
\mbox{N}\Uparrow(\mbox{N}\Uparrow\mbox{S})\qquad$\label{invalid-seq}}
is not derivable, whereas the corresponding ``semantic'' sequent
(up to permutation)
\enumsentence{$q:(e\rightarrow t)\rightarrow t\Rightarrow$\\
\hspace*{2em} $\lam{R}\lam{P}q(\lam{x}R(P)(x)):((e\rightarrow
t)\rightarrow (e \rightarrow t))\rightarrow ((e\rightarrow
t)\rightarrow t)$\label{sem-seq}}
is derivable in the undirected Lambek calculus. Sequent
(\ref{sem-seq}) will in particular raise (\ref{e-u-typ}) to a function
that, applied to (\ref{a-conv-type}), produces (\ref{e-u-a-c-typ}), as
required.

Furthermore,
the solution proposed by \namecite{Morrill:type-logical} to make the
scope calculus complete is to restrict the intended interpretation of
$\Uparrow$ so that (\ref{invalid-seq}) is not valid. Thus, {\em
contra} \namecite{Carpenter:quant-scope}, Morrill's logically more satisfying
account of $\Uparrow$ is not a step towards making reading
(\ref{conv-2}) available.

We now give the derivation of the interpretation (\ref{conv-2}) in our
framework. The f-structure for (\ref{conv}) is:
\enumsentence{\evnup{
\fd{$f$:\fdand{\feat{\pred}{\sc `seek'}
           \feat{\subj}{$g$:\fdand{\feat{\pred}{\sc `Bill'}}}
           \feat{\obj}{$h$:\fdand{\feat{\spec}{\sc `a'}
                                  \feat{\pred}{\sc `conversation'}
				  \feat{\obl\downlett{WITH}}{$i$:
		\fdand{\feat{\spec}{\sc `every'}
		       \feat{\pred}{\sc `unicorn'}}}}}}}}\label{conv-fs}}
The two formulas \BF{Bill-seeks} and \BF{every-unicorn} can be derived
as described before:
$$\begin{array}[t]{ll}
\BF{Bill-seeks}\colon & \forall Y. (\forall s, p. (\forall X.
   h_\sigma \means X \linimp s \means pX)
   \linimp s \means Y({\intn}p))\\
   & \qquad \linimp f_\sigma \means \IT{seek}(\IT{Bill}, {\intn}Y)\\[2ex]
\BF{every-unicorn}\colon\ & \forall G, S.
\/(\forall x.  i_{\sigma} \means x \linimp G \means Sx)\\
   & \qquad \linimp G \means \IT{every}(u, \IT{unicorn}(u), Su)
\end{array}$$
As explained in more detail in Dalrymple, Lamping, Pereira, and
Saraswat (1993), \nocite{DLPS:quant-lfg} the remaining lexical premises
for (\ref{conv-fs}) are:
$$\begin{array}[t]{lr@{}l}
\BF{a}\colon &
\forall H, R, T. & (
(\forall x. (h_{\sigma} \var) \means x \linimp (h_{\sigma} \restr) \means Rx)\\
&& \llap{$\otimes$} \/(\forall x.  h_{\sigma} \means x \linimp H \means Tx))\\
&& \qquad \linimp H \means \IT{a}(z, Rz, Tz)\\[2ex]
\BF{conv-with}\colon &
\forall Z, X.\ &
(h_{\sigma} \var) \means Z \otimes i \means X\\
&&  \qquad \linimp  (h_{\sigma} \restr) \means \IT{conv-with}(Z, X)
\end{array}$$
{}From these premises we immediately derive
\[
\begin{array}[t]{l}
\forall X, H, T. i_{\sigma} \means X \otimes (\forall x.  h_{\sigma} \means x
\linimp H \means Tx))\\
\qquad \linimp H \means \IT{a}(z, \IT{conv-with}(z,X), Tz)
\end{array}\]
which can be rewritten as:
\enumsentence{$\begin{array}[t]{ll}
\forall H, T.
\lefteqn{(\forall x.  h_{\sigma} \means x \linimp H \means Tx)
\linimp} \\
& \All{X}(i_{\sigma} \means X \linimp H \means \IT{a}(z, \IT{conv-with}(z,X),
Tz))
\end{array}$\label{a-conv-with}}
With the substitutions
\[X  \mapsto x,G \mapsto H, S \mapsto \lambda v . \IT{a}(z,
\IT{conv-with}(z,v), Tz))
\]
formula \pex{a-conv-with} can be combined with
\BF{every-unicorn} to yield the required quantifier-type formula:
\enumsentence{$\begin{array}[t]{ll}
\lefteqn{\forall H, T. (\forall x.  h_{\sigma} \means x \linimp H \means Tx)
\linimp} \\
& H \means \IT{every}(u, \IT{unicorn}(u), \IT{a}(z, \IT{conv-with}(z,u), Tz))
\end{array}$\label{a-conv-w-e-un}}
Using substitutions
\[\begin{array}{rcl}
H & \mapsto & s \\
T & \mapsto & p \\
Y & \mapsto &\lambda R.
\IT{every}(u,\IT{unicorn}(u),\IT{a}(z,\IT{conv-with}(z,u),[{\extn}R](z))))
\end{array}
\]
and modus ponens, we then combine \pex{a-conv-w-e-un} with
\BF{Bill-seeks} to obtain the desired final result:
$$
f_\sigma \means
\IT{seek}(\IT{Bill}, {\intn}\lambda R.
\IT{every}(u,\IT{unicorn}\/(u),a\/(z,\IT{conv-with}(z,u),[{\extn}R](z))))
$$

We see thus that our more flexible connection between syntax and
semantics permits the full range of type flexibility provided
categorial {\em semantics} without losing the rigorous connection to
syntax. In contrast, current categorial accounts of the
syntax-semantics interface do not appear to offer the needed
flexibility when syntactic and semantic composition are more indirectly
connected, as in the present case.

\section{Conclusion}

We have shown that our deductive framework allows us to predict the
correct set of readings for intensional verbs with quantified and
nonquantified direct objects if we make a single assumption: that
intensional verbs require a quantified direct object. This assumption
is, of course, the starting point of the standard Montagovian
treatment of intensional verbs. But that treatment depends on the
additional machinery of quantifying in to generate {\em de re}
readings of quantified direct objects, and that of explicit type
raising to accommodate unquantified direct objects.  In our approach
those problems are handled directly by the deductive apparatus without
further stipulation.

These results, as well as our previous work on quantifier scope,
suggest the advantages of a generalized form of compositionality in
which the semantic contributions of phrases are represented by logical
formulas rather than by functional abstractions as in traditional
compositionality. The fixed application order and fixed type
requirements of lambda terms are just too restrictive when it comes to
encoding the freer order of information presentation in natural
language.  In this observation, our treatment is closely related to
systems of syntactic and semantic type assignment based on the Lambek
calculus and its variants. However, we differ from those categorial
approaches in providing an explicit link between functional structures
and semantic derivations that does not depend on linear order and
constituency in syntax to keep track of predicate-argument relations.
Thus we avoid the need to force both syntax and semantics into an
uncomfortably tight categorial embrace.

\section*{Acknowledgments}
We thank David Israel, Michael Moortgat and Stanley Peters for
discussions on the subject of this paper.

\end{document}